\documentclass[review]{elsarticle}
\usepackage[top=1in,bottom=1in,left=1in,right=1in]{geometry}
\usepackage{latexsym, graphicx, epsfig, amsmath, amssymb, amsfonts}
\usepackage{amsmath, mathrsfs, mathtools, amsthm}
\usepackage{amssymb}
\usepackage{graphicx}
\usepackage{subfig}
\usepackage{tabu}
\usepackage{algorithm}
\usepackage{algpseudocode}
\usepackage{adjustbox}
\usepackage{bbold}
\usepackage{bm}
\usepackage{textcomp}
\usepackage[thinc]{esdiff}
\usepackage[dvipsnames]{xcolor}
\usepackage{comment}
\usepackage{soul}
\usepackage{multirow}
\usepackage[]{mdframed}
\biboptions{sort&compress}
\newtheorem{theorem}{Theorem}

\def\Omegatheta{{\Omega^\theta}}
\def\Omegathetai{{\Omega^{\theta_i}}}

\def\refd{0}
\def\Omegaref{{\Omega^\refd}}
\def\vv0thetak{v^\refd_{k}}
\def\u0theta{u^\refd_{\theta}(\cdot;\mathbf{v}^\refd_\theta)}
\def\bfvtheta{\mathbf{v}^\theta}
\def\F0{{\mathcal{F}_{\refd}}}
\def\x{x}

\newcommand\norm[1]{\left\lVert#1\right\rVert}

\begin{document}
\begin{frontmatter}

\title{DIMON: Learning Solution Operators of Partial Differential Equations on a Diffeomorphic Family of Domains}

\author{Minglang Yin\textsuperscript{a}}
\author{Nicolas Charon\textsuperscript{b}}
\author{Ryan Brody\textsuperscript{a,e}}
\author{Lu Lu\textsuperscript{c}}
\author{Natalia Trayanova\textsuperscript{a,d,e}\corref{cor}}
\author{Mauro Maggioni\textsuperscript{d,f}\corref{cor}}

\cortext[cor]{Corresponding authors: ntrayanova@jhu.edu and mauromaggionijhu@icloud.com}
\address[a]{Department of Biomedical Engineering, Johns Hopkins University, Baltimore, MD}
\address[b]{Department of Mathematics, University of Houston, Houston, TX}
\address[c]{Department of Statistics and Data Science, Yale University, New Haven, CT}
\address[d]{Department of Applied Mathematics and Statistics, Johns Hopkins University, Baltimore, MD}
\address[e]{School of Medicine, Johns Hopkins University, Baltimore, MD}
\address[f]{Department of Mathematics, Johns Hopkins University, Baltimore, MD}

\begin{abstract}
The solution of a PDE over varying initial/boundary conditions on multiple domains is needed in a wide variety of applications, but it is computationally expensive if the solution is computed de novo whenever the initial/boundary conditions of the domain change. 
We introduce a general operator learning framework, called DIffeomorphic Mapping Operator learNing (DIMON) to learn approximate PDE solutions over a family of domains $\{\Omegatheta\}_\theta$, that learns the map from initial/boundary conditions and domain $\Omega_\theta$ to the solution of the PDE, or to specified functionals thereof.
DIMON is based on transporting a given problem (initial/boundary conditions and domain $\Omegatheta$) to a problem on a reference domain $\Omegaref$, where training data from multiple problems is used to learn the map to the solution on $\Omegaref$, which is then re-mapped to the original domain $\Omegatheta$.  
We consider several problems to demonstrate the performance of the framework in learning both static and time-dependent PDEs on non-rigid geometries; these include solving the Laplace equation, reaction-diffusion equations, and a multiscale PDE that characterizes the electrical propagation on the left ventricle. This work paves the way toward the fast prediction of PDE solutions on a family of domains and the application of neural operators in engineering and precision medicine. 
\end{abstract}

\end{frontmatter}
\textbf{keywords:} Deep Learning, Neural Operator, Partial Differential Equations, Parametric Domains\\
\textbf{author contributions}: M.Y., N.C., M.M, and N.T. designed research; M.Y., N.C., and R.B. performed research; M.Y., L.L., and M.M. analyzed data; M.Y., N.C., L.L., and M.M. contributed to the theoretical results; M.Y., N.C., R.B., L.L., M.M., and N.T wrote the paper.\\
\textbf{Autho Declaration:} Declare no competing interests.\\
\textbf{Corresponding authors:} To whom correspondence should be addressed. E-mail: Mauro Maggioni: mauromaggionijhu@icloud.com and Natalia Trayanova: ntrayanova@jhu.edu

\newpage

\section{Significance} \label{sec:sig}

Recent advancements in deep learning have been used to expedite the estimation of partial differential equation (PDE) solutions. However, there still lacks a generic framework that enables learning across a family of domains, hindering the expansion of deep learning in PDE-based engineering and medical problems. In this study, we tackle this challenge by developing a generic framework that combines diffeomorphism and deep learning for fast prediction of PDE solutions on a family of domains without retraining. The proposed framework paves the way for the utilization of deep learning for engineering or medical applications, such as design and optimization.

\section{Introduction}\label{sec:intro}

The solution of partial differential equations (PDEs) for multiple initial and boundary conditions, and over families of domains or shapes are essential in a wide variety of disciplines, including applied mathematics, engineering, or medical sciences, with applications such as topology/design optimization and clinical prognostication being particularly demanding in terms of requiring the solution of PDEs on multiple domains and with many different initial and boundary conditions. Independent approximations of the solution in each instance using standard numerical methods, although effective, can be computationally costly, and techniques that aim at improving efficiency by exploiting relationships between solutions over different shapes and initial/boundary conditions have received significant attention in the research community.

Within the emerging field of scientific machine learning, many approaches have been proposed to approximate the solutions of PDEs by deep learning models. The idea of solving PDEs with neural networks (NN) can be traced back at least to 90s (see e.g.~\cite{lagaris1998artificial,chen1995universal}), and has received much attention in the last decade due to the advance in computational ability for training deep neural networks (DNN) (see e.g.~\cite{karniadakis2021physics,raissi2019physics,yu2018deep,lu2021deepxde,cai2021physics}).In particular, using deep learning to learn the PDE solution operator, termed neural operator, is of particular interest as it enables approximating PDE solutions with varying boundary/initial conditions, and over parametric families of PDEs (e.g. modeling local material properties) ~\cite{lu2021learning,li2020fourier}. 

However, the need to estimating PDE solutions on various domains or shapes is yet to be fully addressed. Current frameworks for neural operators are formulated on a domain with fixed boundaries, and even slight variations in the shape of the learning domain require retraining the neural network. Previous attempts to extend the learning ability on various geometries adopted approaches based on level-sets~\cite{li2023geometry}, convolutional neural networks~\cite{he2023novel}, among others~\cite{li2022geoFNO}, although with rather significant limitations in each of these studies. The challenge of learning operators on various geometries emanates from learning functional mappings, not on a fixed, but a family of Banach spaces over various domains, yet none of the previous studies provided a comprehensive and generic formulation that reconciles the learning problems with their supporting approximation theorem.

Herein, we propose a theoretically sound and computationally accurate operator learning framework, termed DIffeomorphic Mapping Operator learNing (DIMON), that extends operator learning to a family of diffeomorphic domains. DIMON combines neural operators with diffeomorphic mappings between domains/shapes, formulating the learning problem on a unified reference domain (``template") $\Omegaref$, but with respect to a suitably modified PDE operator. PDE solutions along with input functions on a family of diffeomorphically equivalent shapes are each mapped onto the reference domain $\Omegaref$ via a diffeomorphism. This map can also be interpreted as a change of variables in the PDE on the original shape, yielding a corresponding PDE on $\Omegaref$, parameterized by the diffeomorphism. A neural operator is trained to learn the latent solution operator of this parameterized PDE on the reference domain $\Omegaref$, where the parameter depends on the mapping from the original domain to the reference domain. The original operator is equivalent to the composition of the learned latent operator with the inverse mapping (pushforward). Learning this parameterized family of PDE operators using neural operators then yields a way of solving PDEs on the family of domains. 

Although the DIMON framework requires no restriction on the structure of the neural operators, in order to demonstrate its utility we extend the universal approximation theorem in~\cite{jin2022mionet} and show that, in practice, the latent operator on $\Omegaref$ can be approximated by adopting a multiple-input operator (MIONet). In addition, DIMON is flexible with the choice of the diffeomorphic mapping algorithms and reference domain $\Omegaref$, which facilitates the adoption of existing mapping algorithms under our framework~\cite{ovsjanikov2012functional,vercauteren2009diffeomorphic}. We demonstrate the ability of our framework with three examples, including learning static and dynamic PDE problems on both parameterized and non-parameterized synthetic domains. In particular, the last example approximates functionals of PDE solutions (in fact, PDEs couple with a rather high-dimensional ODE system) for determining electrical wave propagation in patient-specific geometries of the heart, demonstrating its utility in learning on realistic shapes and realistic dynamics in precision medicine applications.

\section{Results}
\label{sec:res}

\subsection{Problem Formulation}
\label{subsec:prob_formul} 
In this section, we present the formulation of DIMON for learning operators on a family of diffeomorphic domains, along with the unfolding of how a pedagogical example problem is set up and solved. 

\noindent{\em{Setup}}. Consider a family $\{\Omegatheta\}_{\theta\in\Theta}$ of bounded, open domains with Lipschitz boundaries in $\mathbb{R}^{d}$, with $\Theta$ a compact subset of $\mathbb{R}^p$. Assume that for each shape parameter $\theta\in\Theta$ there is a $\mathcal{C}^2$ embedding $\varphi_\theta$ from the ``template'' $\Omegaref$ to $\mathbb{R}^{d}$, which induces a $C^2$ diffeomorphism (i.e. a bijective $\mathcal{C}^2$ map with $\mathcal{C}^2$ inverse) between $\Omegaref$ and $\Omegatheta$. We further assume that the map $\Theta\rightarrow \mathcal{C}^2(\Omegaref,\mathbb{R}^{d})$, defined by $\theta\mapsto\varphi_\theta$ is continuous, when $\mathcal{C}^2$ is endowed with the usual $\mathcal{C}^2$-norm that makes it a Banach space.
Assume that for each $\theta \in \Theta$ there is a solution operator $\mathcal{G}_{\theta}:X_1(\Omegatheta)\times\dots X_m(\Omegatheta)\rightarrow Y(\Omegatheta)$ mapping functions $v^{\theta}_{1},\dots,v^\theta_m$ on $\Omegatheta$ or $\partial\Omegatheta$, representing a set of initial/boundary conditions, to the solution $u^{\theta}(\cdot;\bfvtheta)$ on $\Omegatheta$ of a fixed PDE, where $\bfvtheta:=(v^\theta_1,\dots,v^\theta_m)$. 

\begin{mdframed} 
\textbf{Example:} Let us consider the problem of solving the Laplace equation on a family of bounded open domains $\{\Omegatheta\}_{\theta \in \Theta}$ of $\mathbb{R}^{2}$ with Lipschitz regular boundaries, where $\Theta$ is a compact subset of $\mathbb{R}^p$. 
We assume that $\Omegatheta$ is, for each $\theta\in\Theta$, diffeomorphic to a fixed domain $\Omegaref\subset \mathbb{R}^2$, which we call the reference domain. For example, if $\Omegatheta$ is a simply connected domain (different from the whole $\mathbb{R}^d$) for each $\theta$, $\Omegaref$ may be chosen to the open unit disk $\{x\in\mathbb{R}^d\,:\,||x||_2<1\}$, and $\varphi_\theta$ smooth (in fact, holomorphic if $\mathbb{R}^2$ is identified with $\mathbb{C}$) bijection of $\Omegatheta$ onto $\Omegaref$ exists by the Riemann mapping theorem.

For each $\theta\in\Theta$, we are interested in obtaining the solution of the PDE
\begin{equation}
\label{eq:Laplace_PDE}
\begin{cases}
    \Delta u &= 0 \qquad \text{in }\Omegatheta, \\
    g&=v_1^\theta \qquad \text{on }\partial\Omegatheta.
\end{cases}
\end{equation}
for $v_1^\theta$ a function on $\partial \Omegatheta$ with $v_1^\theta \in H^{\frac{1}{2}}(\partial \Omegatheta)$, and with the boundary condition intended in the trace sense. In this setting, it is known that there exists a unique weak solution $u^{\theta}(\cdot;v_1^\theta) \in H^1(\Omegatheta)$ to this boundary problem, c.f. for instance \cite{gilbarg1977elliptic} or \cite{jerison1995inhomogeneous}. We are interested in learning the solution operator $\mathcal{G}_\theta(v_1^\theta):=u^{\theta}(\cdot;v_1^\theta)$ for this Laplace equation over the various domains $\Omegatheta$ with boundary conditions $v_1^\theta$.
\end{mdframed}

We may represent this family $\{\mathcal{G}_{\theta}\}_{\theta\in\Theta}$ of operators defined on functions on different domains by an operator $\F0$ dependent on the shape parameter $\theta$ and on functions that are all defined on the reference domain $\Omegaref$, by suitable compositions with the diffeomorphisms $\varphi_\theta$ and $\varphi_\theta^{-1}$, as follows:
\begin{align}
    u^{\theta}(\cdot;\mathbf{v}^\theta)
    = \mathcal{G}_{\theta}(\underbrace{v^{\theta}_{1},...,v^{\theta}_{m}}_{\bfvtheta})
    = \F0(\theta,     \underbrace{v^\theta_1\circ\varphi_{\theta},\dots,v^\theta_m\circ\varphi_{\theta}}_{\bfvtheta \circ \varphi_{\theta}})\circ \varphi_{\theta}^{-1}
    = \F0(\theta, \underbrace{v_{\theta,1}^\refd,\dots,v_{\theta,m}^\refd}_{\mathbf{v}^{0}_{\theta}})\circ \varphi_{\theta}^{-1} 
    = \u0theta\circ\varphi_{\theta}^{-1}\, ,
\label{e:defF0}
\end{align}
where the second equality defines $\F0$, the third equality defines $\vv0thetak:=v_k^\theta\circ\varphi_{\theta}$, and the fourth equality shows that $\u0theta:=u^{\theta}(\cdot;\mathbf{v}^\theta)\circ\varphi_{\theta}$.The ``latent'' operator $\F0: \Theta \times X_{1}(\Omegaref)\times \cdots \times X_{m}(\Omegaref) \rightarrow Y(\Omegaref)$ maps $\theta\in \Theta$ and a set of functions $v^{\refd}_{k}$ in some Banach Spaces $(X_{k}(\Omegaref), ||\cdot||_{X_{k}(\Omegaref)})$, for $k=1,...,m$, of functions on $\Omegaref$ or $\partial\Omegaref$, to a target function $u^\refd_{\theta,\mathbf{v}^\refd}$ in a Banach Space $(Y(\Omegaref),||\cdot||_{Y(\Omegaref)})$ of functions over $\Omegaref$. As above, we used the vector notation $\mathbf{v}^\refd$ to denote $(v^\refd_1,\dots,v^\refd_m)$. One can view $\F0$ as the operator that outputs the pull back of a solution of a PDE on $\Omegatheta$ to the reference domain $\Omegaref$ via the diffeomorphism $\varphi_{\theta}$; this is analogous to the \textit{Lagrangian} form of the solution operator in contrast with the \textit{Eulerian} form $\mathcal{G}_{\theta}$ defined on the varying domain using the terminology of \cite{henry2005perturbation}.  

\begin{mdframed}
\textbf{Example:}
Going back to the example of the Laplace equation, the operator $\F0$ to approximate maps a shape parameter $\theta$ and boundary function $v_1^\theta$ to the (weak) solution of \eqref{eq:Laplace_PDE} pulled back on the template domain $\Omegaref$. Formally, denoting $u^\refd_{\theta}(\cdot;v^\refd_\theta) = \F0(\theta,v^\refd_\theta)$, where $v^\refd_\theta = v_1^\theta \circ \varphi_\theta$, $u^\refd_{\theta}(\cdot;v^\refd_\theta)$ can be equivalently viewed as the solution of a uniformly elliptic Dirichlet problem on $\Omegaref$, namely 
\begin{equation}
\label{eq:Laplace_PDE_can_domain}
\begin{cases}
    \varphi_\theta^* \, \Delta \, (\varphi_\theta^{-1})^* \, u^\refd_{\theta}(\cdot;v^\refd_\theta) &= 0 \qquad \text{in }\Omegaref \\
    g&=v^\refd_\theta \qquad \text{on } \partial\Omegaref
\end{cases}\,,
\end{equation}
in which $\varphi_\theta^*$ denotes the operator mapping an arbitrary function $h$ on $\Omega_\theta$ to the function $h \circ \varphi_\theta$ on $\Omegaref$. Note that the specific coefficients of this linear second-order elliptic operator $L_\theta = \varphi_\theta^* \, \Delta \, (\varphi_\theta^{-1})^*$ could be written explicitly based on the first and second order derivatives of $\varphi_\theta$, and are in particular continuous functions on $\Omega^0$, c.f. \ref{append:laplace_case} for more details.
\end{mdframed}

\noindent{\em{Learning problem}}. Given $N$ observations $\{(\Omegathetai,v_1^{\theta_i} \dots,v_m^{\theta_i},\mathcal{G}_{\theta_i}(v^{\theta_i}_{1},...,v^{\theta_i}_{m}))\}_{i=1}^N$ and a diffeomorphism $\varphi_\theta$, where $\theta_i$ and $v_k^{\theta_i}$ are all independently sampled at random from a probability measure on $\Theta$ and on compact subsets of $X_k$, respectively, we aim at constructing an estimator of $\F0$ defined in \eqref{e:defF0}.

\begin{figure}
    \centering
	\includegraphics[width=1.0\textwidth]{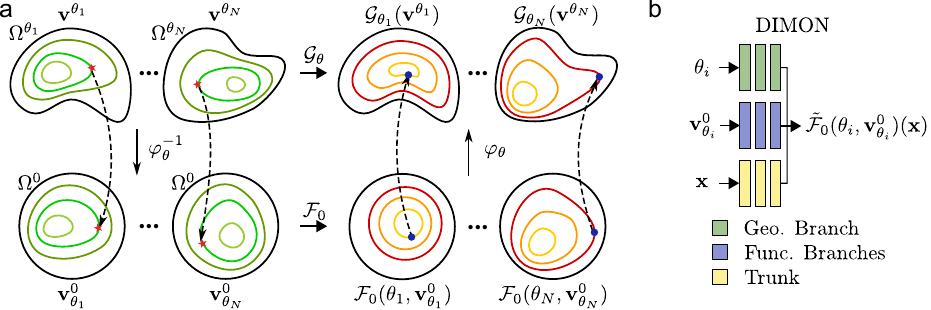} 
    \caption{\textbf{Schematics of DIMON for operator learning on a family of diffeomorphic domains.} \textbf{a}: A partial differential operator $\mathcal{G}_{\theta}$ is posed on $\Omegathetai$ together with functions $\mathbf{v}^{\theta_i} (i=1,\dots, N)$ as input. A map $\varphi_{\theta}$, parameterized by the shape parameter $\theta$, transports $\mathbf{v}^{0}_{\theta}$ posed on reference domain $\Omegaref$ onto $\Omegatheta$ and identifies point correspondences across all $\Omegatheta$ and $\Omegaref$ (indicated by red marks and blue dots). There exists a latent operator $\F0$ such that $\mathcal{G}_{\theta}(\bfvtheta) = \F0(\theta, \mathbf{v}^0_\theta)\circ \varphi_{\theta}^{-1}$.  \textbf{b}: The operator $\F0$ is approximated by a neural operator $\tilde{\mathcal{F}}_{0}$: a shape parameter $\theta$ (typically estimated from data) and the transported function $\mathbf{v}^{0}_{\theta}$ are the inputs of branch networks; the coordinates of the reference domain $\x $ are the inputs of the trunk network.}
    \label{fig:schematics}
\end{figure}
We choose the estimator in the form of a neural operator $\tilde{\mathcal{F}}_0$, which takes as inputs the shape parameter $\theta$, the transported input functions $\mathbf{v}^0_\theta$, and outputs an approximation to $\F0(\theta,\mathbf{v}^\refd_\theta)$. We visualize the framework in Fig.~\ref{fig:schematics}a and the high-level architecture of the neural operator in Fig.~\ref{fig:schematics}b; we discuss more details of the neural operator and its approximation ability in Sec.~\ref{subsec:theorem}. 

The framework discussed above extends immediately to at least the following settings:
\begin{itemize} 
\item time-dependent operators, by concatenating time $t$ to $\x $ (see Example 2);
\item operators that involve vector- or tensor-valued functions, by replacing the pullback operation on scalar functions with an appropriate action of diffeomorphisms on the vector and tensor fields. For example, the action of $\varphi_{\theta}$ on a vector field $v:\Omegaref \rightarrow \mathbb{R}^d$ is the vector field defined for all $x_\theta=\varphi_{\theta}(x) \in \Omegatheta$ by: $J\varphi_{\theta} \cdot v(x)$ where $J\varphi_{\theta}$ denotes the Jacobian matrix of the transformation. 
This action can be further extended to covariant/contravariant tensor fields of any type.
\end{itemize}

\subsection{Universal Approximation Theorem on a Diffeomorphic Family of Domains}
\label{subsec:theorem}

In this section, we extend the applicability of the universal approximation theorem (UAT) presented in Jin et al.~\cite{jin2022mionet} on learning parametric domains for multi-input operators under DIMON framework. In short, the UAT in \cite{jin2022mionet} guarantees the approximation ability of a multi-branch DeepONet for nonlinear and continuous operators with the assumption of fixed domains. Transporting functions defined on various domains via $\varphi_{\theta}$ onto a reference domain with a fixed boundary naturally aligns well with the assumption of UAT, hence enabling the ability to learn on various domains. To differentiate the mapping $\varphi$ and the shape of $\Omegatheta$, we define a shape parameter $\theta \in \Theta \subset X_{0}$ in a Banach space $X_{0}$ and introduce the branch for encoding geometric information, referred to as Geo. branch. Therefore, the adapted UAT can be written as:

\begin{theorem}[Adapted from \cite{jin2022mionet}] 
\label{thm:universal_approx}
Assume that $X_{0}, X_{1},\dots,X_{m}, Y$ are Banach Spaces, $\Theta \subset X_{0}, K_{i}\subset X_{i}$ ($i=1,...,m$) are compact sets, and $X_{i}$ ($i=0,...,m$) has a Schauder basis. If $\F0: \Theta \times K_{1}\times \cdots \times K_{m}\rightarrow Y$ is a continuous operator, then for any $\epsilon>0$, there exists a neural network $\tilde{\F0}$ with the form given in Eq.~\eqref{equ:no_general}, such that 
$$\norm{\F0-\tilde{\F0}}_{C(\Theta \times K_1\times\cdots\times K_m,Y)}<\epsilon\,,$$
where the norm $C$ is the sup norm.
\end{theorem}
Despite its generality, the use of Theorem \ref{thm:universal_approx} for the approximation of solution operators of PDEs typically requires extra analysis, in particular to ensure that the assumptions on $\F0$ are satisfied, namely the well-posedness and continuity with respect to the domain geometry as well as initial/boundary conditions: this can be non-trivial for many PDEs. While it is out of the scope of the present paper to discuss this in a systematic fashion, we will detail below and in the appendix the particular case of the Laplace equation introduced above, under suitable regularity assumptions. 

\begin{mdframed}
\textbf{Example:}
Going back to the example of the Laplace equation of Sec. \ref{subsec:prob_formul}, the key issue to apply Theorem \ref{thm:universal_approx} in this case is to determine adequate functional spaces for the solution and boundary function under which $u^\refd_{\theta}(\cdot;v^\refd_\theta)$ is first well-defined but also continuous with respect to $v^0_\theta$ and $\theta$. The existence and uniqueness of weak solutions in $H^1(\Omegaref)$ for smooth domains and $v^0_\theta \in H^{1/2}(\partial \Omegaref)$ follow from classical results on elliptic PDEs (see e.g. \cite{gilbarg1977elliptic} Chap.8), and have been extended to Lipschitz regular domains in \cite{jerison1995inhomogeneous}. Regularity estimates for this class of PDEs also guarantee the continuity of the solution with respect to the boundary function $v^0_\theta$. However, to the authors' knowledge, regularity with respect to domain deformations (and thus $\theta$) is much less known in such a general setting. 
Based on the continuity of the mapping $\theta \mapsto \varphi_{\theta}$ from $\Theta$ to $\mathcal{C}^2(\Omegaref,\mathbb{R}^d)$ and by adapting the ideas presented in \cite{henry2005perturbation}, one can recover the solution's continuity with respect to $\theta$ albeit under the stronger assumption that $\Omega^0$ has a $C^2$ boundary. For the sake of concision, we refer to \ref{append:laplace_case} for some additional details. With such assumptions, the conclusion of Theorem \ref{thm:universal_approx} holds and it is thus possible to approximate the solution operator up to any precision using the operator learning framework developed in this paper. 
\end{mdframed}

\subsection{Algorithms and practical considerations}
In practice, the domain $\Omegatheta$ may be available only in discretized form, and numerical methods are used to approximate the solution of a PDE on the domain of $\Omegatheta$. For all the examples presented in this work, we adopt a FEM with a piece-wise linear basis to generate a training set of PDE solutions, consistent with the assumption $u^{\theta}\in H^1(\Omegatheta)$ on the weak solution. One could adopt numerical methods with higher-order polynomials, such as spectral element methods~\cite{karniadakis2005spectral,patera1984spectral} or $hp$-FEM~\cite{babuvska1990p}, in order to approximate smoother PDE solutions with higher accuracy; note that in this case the differentiability of $\varphi_{\theta}$ needs to be sufficiently high to preserve the smoothness solution after composition.

\subsubsection{Choosing the reference domain $\Omegaref$}

In general, there is no canonical choice for $\Omegaref$, or for the diffeomorphisms between $\Omegaref$ and $\Omega^\theta$. Often the domains are unparameterized, given for instance meshes with no explicit point correspondences between them. For such unparameterized domains, a practical choice is then to set a reference domain that has a simple shape and is diffeomorphically equivalent to all the $\Omegatheta$'s. For instance, we choose a hollow semi-sphere as $\Omegaref$ as a simple shape visually close to the left ventricle in Example 3. In general, one may select one particular $\Omega^{\theta_i}$, with $\theta_i \in \Theta$, as reference, or estimate a template shape, constrained to lie in $\{\Omega^\theta\}$, from a given dataset of sample domains using one of the many possible approaches developed for this purpose including, among others, \cite{avants2004geodesic,joshi2004unbiased,ma2010bayesian,cury2017analysis,hartman2023elastic}. Although one can measure the similarity/distance between domains using a metric~\cite{miller2002metrics}, i.e., Chamfer distance~\cite{fan2017point,park2019deepsdf}, Earth Movers' distance~\cite{rock2015completing,rubner2000earth}, or Riemannian metrics on shape spaces~\cite{beg2005computing,grenander1998computational,bauer2012sobolev}, this is not a prerequisite for successfully adopting this learning framework. This flexibility in choosing the reference domain facilitates a wide set of applications.

\subsubsection{Estimating and encoding the domain geometry with the shape parameter}\label{sssec:pc}
Once a reference domain $\Omegaref$ is chosen, the diffeomorphisms $\varphi_{\theta}$ that map $\Omegaref$ to the domains $\Omegatheta$ need to be constructed. In the case of a family of parameterized shapes, as in the examples of Section \ref{subsec:lap} below, the $\varphi_{\theta}$ can be constructed naturally by composing parameterizations, which also yields point correspondences, see Sec.~\ref{subsec:dom_gen}. 

For unparameterized data however, such as the meshes in the example of Section \ref{subsec:rd} and \ref{subsec:LV}, point correspondences are not a priori available and the mesh structure itself may not even be consistent from one data sample to another. In such cases, the maps $\varphi_{\theta}$ are not known and need to be approximated for each domain. A variety of models can be used for this purpose relying, for example, on maps that are area-preserving~\cite{desbrun2002intrinsic}, {quasi-conformal} ~\cite{praun2003spherical,choi2022recent} or {quasi-isometric}~\cite{sun2008quasi}, and which can be estimated for instance via the versatile functional maps framework of~\cite{ovsjanikov2012functional,ovsjanikov2016computing}. Yet, in some applications, one may need a larger class of possible deformations, especially if larger distortions occur between different domains. For that reason, in the present work, we choose to rather rely on the diffeomorphic registration setting introduced in~\cite{beg2005computing}, known as Large Deformation Diffeomorphic Metric Mapping (LDDMM). This model has several notable advantages when it comes to the applications we consider here. First, it generates transformations as flows of smooth vector fields that result in smooth diffeomorphisms of the whole ambient space $\mathbb{R}^d$, which is consistent with the theoretical setting developed in the previous section. From these estimated deformations, one can easily recover approximate point correspondences between $\Omegaref$ and the other domains, as well as map functions but also vector fields or tensor fields from one domain to another. Another interesting feature of LDDMM is the fact that it is connected to a Riemannian metric on the diffeomorphism group \cite{younes2019shapes}. Thus, any optimal deformation path in that model can be interpreted as a geodesic and, as such, be represented compactly via some initial "momentum" field defined on $\Omegaref$. The Riemannian setting further enables the extension of many statistical techniques to the space of deformations, including methods such as principle component analysis (PCA) for dimensionality reduction, as we discuss next.

In the perspective of learning the solution operator over a family of diffeomorphic domains, we also seek a representation of the deformations $\varphi_\theta$ that can be effectively used as input for the operator learning neural network. The estimator of the map $\varphi_\theta$ generally leads to high-dimensional objects, either a large set of point correspondences between the mesh of $\Omegaref$ and $\Omega^\theta$ or as a set of momentum vectors attached to vertices of $\Omegaref$ in the case of the LDDMM framework described above. In order to obtain an informative representation encoded by lower dimensional shape parameter $\tilde{\theta}$, we apply dimensionality reduction (herein PCA, see Sec.~\ref{subsec:pca} for additional details). A first approach is to apply standard PCA to the set of displacement vectors $\{x-\varphi_\theta(x)\}_{x\in\mathrm{lndmrk}(\Omegaref)}$, where $\mathrm{lndmrk}(\Omegaref)$ is a fixed set of landmark points in $\Omegaref$. Alternatively, when deformations are estimated from the LDDMM model, we instead leverage the geodesic PCA method \cite{fletcher2004principal} on either the momentum or the deformation representation. In either case, projection onto the first $p'$ principal components leads to our estimated low-dimensional representation $\tilde{\theta}$ of the shape parameter $\theta$ for each $\Omegatheta$, and gives a more compact representation of $\varphi_\theta$. Notice that we only adopt reduced-order PCA coefficients to represent the shape of $\Omegatheta$ for network training, but will still rely on the originally (non-reduced) estimated mappings $\varphi_\theta$ to transport the estimated solution on $\Omegaref$ back to $\Omegatheta$.

\subsection{Algorithms}

We summarize our procedure in Algorithm ~\ref{alg:1}. The computational complexity is dictated by the complexity of solving the PDE on $\Omega^\theta$ and the training for the neural network operator. 

\begin{algorithm}[H]
\caption{The algorithm of learning solution operators on parametric domains.}\label{alg:1}
    \begin{algorithmic}
 \Statex \textbf{Data:} $\{(\Omegathetai,\mathbf{v}^{\theta_i},u^{\theta_i}(\cdot;\mathbf{v}^{\theta_i})\}_{i=1}^N$: numerical PDE solution $u^{\theta_i}(\cdot;\mathbf{v}^{\theta_i})$, with input functions $\mathbf{v}^{\theta_i}$, on $\Omegathetai$
 \Statex 1. set a reference domain $\Omegaref$ and compute point correspondences between $\Omegathetai$ and $\Omegaref$\;
 \Statex 2. calculate the reduced-order shape parameters $\tilde{\theta}_i$, using for example PCA\;
 \Statex 3. pushforward $\mathbf{v}^{\theta_i}$ and $u^{\theta_i}$ onto $\Omegaref$: $u^0_{\theta_i}:=u^{\theta_i}\circ\varphi_{\theta_i}$ and $\mathbf{v}^{0}_{\theta_i} := \mathbf{v}^{\theta_i} \circ \varphi_{\theta_i}$\;
 \Statex 4. initialize and train a neural operator $\tilde{\F0}$ to approximate $u^0_{\theta_i}$: 
 \For{epoch = 1:$N_{train}$}
    \State forward propagation $u_{NN,i} :=  \tilde{\mathcal{F}}_{0}(\tilde{\theta},\mathbf{v}^{0}_{\theta_i})$ for all $\theta_i$ in the training dataset \; 
    \State compute the mean square error (MSE) as the loss $\mathcal{L} := \sum_i MSE(u_{NN,i}, u^{0}_{\theta_i})$ \;
    \State update network parameters to minimize $\mathcal{L}$\;
 \EndFor
 \Statex 5. pull network predictions back to $\Omegathetai$ using $ u_{NN,i}\circ \varphi_{\theta_i}^{-1}$.
 \end{algorithmic}
\end{algorithm}

\subsection{Example 1: Solving the Laplace Equation on Parametric 2D Domains}
\label{subsec:lap}

\begin{figure}
    \centering
	\includegraphics[width=1.0\textwidth]{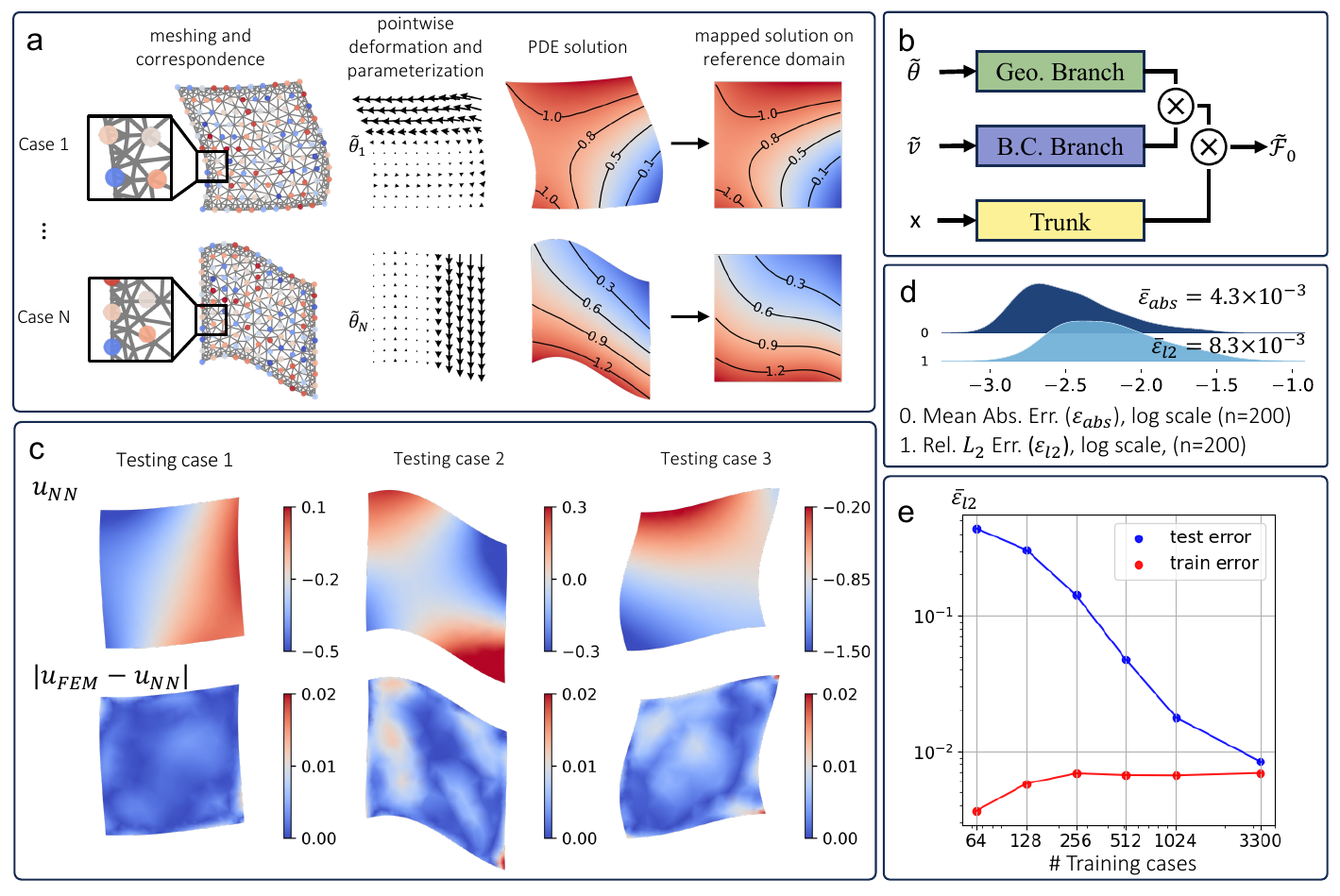} 
    \caption{\textbf{DIMON for learning solution operator of the Laplace equation on parameterized 2D domains.} {\bf a}: Data generation on $N$ parametric 2D domains. Landmark points are matched across 2D domains with different meshing. Shape parameter $\tilde{\theta}\in \mathbb{R}^{p'}$ is calculated by PCA on the deformation field at the landmark points. Here, we adopt the first 15 principal components as $\tilde{\theta}$. The PDE is solved using finite elements and mapped onto the reference domain. {\bf b}: Neural operator structure. Shape parameter $\tilde{\theta}$ and boundary condition $\tilde{v}$ are fed into the geometry (Geo.) and boundary condition (B.C.) branch, respectively. Coordinates in the reference domain $\mathbf{x}$ are the input of the trunk net. {\bf c}: Network predictions ($u_{NN}$) and their absolute errors with numerical solutions ($|u_{FEM} - u_{NN}|$) on three testing domains. {\bf d}: Distributions of mean absolute errors $\varepsilon_{abs}$
    and relative $L_{2}$ error ($\varepsilon_{l2}$) on the testing dataset in log scale. The testing dataset contains $n=200$ cases. {\bf e} The average of $\varepsilon_{l2}$ ($\bar \varepsilon_{l2}$) for the training (red) and testing dataset (blue) is plotted against the number of training cases. 
    }
    \label{fig:laplace}
\end{figure}

\noindent{\bf{PDE problem}}. In this section, we present the network prediction for the pedagogical example of the Laplace equation discussed above. The PDE problem on a domain $\Omegatheta$ is
\begin{align*}
\begin{cases}
    \Delta u &= 0\,,\quad\text{ in } \Omegatheta, \\
    u|_{\partial\Omegatheta} &= v \sim \text{GRF}(0, k_{l})\,,
\end{cases}
\end{align*}
where for the boundary conditions we use the Gaussian Random Field (GRF) with the covariance kernel:
\begin{equation}
    k_{l}(x_{1}, x_{2}) = \exp{(-\norm{x_{1}-x_{2}}}^{2}/2l^{2})\,,
\end{equation}

\noindent{\bf{Data preparation}}. The process of data preparation is illustrated in Fig.~\ref{fig:laplace}a. Following the steps detailed in Method and Materials~\ref{subsec:dom_gen}, we first generate 3,500 samples of the shape parameters $\theta$, create the corresponding domains $\Omegathetai$, which are deformations of the unit square, which itself is set as the reference domain $\Omegaref$. We then impose a boundary condition sampled from a GRF with a length scale $l=0.5$ and solve the governing equation using a finite element solver. At the end of data preparation, we map the PDE solution from $\Omegathetai$ to $\Omegaref$ for network training. 

\noindent{\bf{Construction of $\varphi_\theta$ and $\tilde{\theta}$}}. We obtain a set of point correspondences between $11\times11$ uniformly distributed landmarks on $\Omegaref$ to their images in $\Omegathetai$, using the map $\varphi_{\theta_i}$, which is considered known in this example (see Sec.~\ref{subsec:dom_gen}). We visualize the deformation via vector fields in the second column of Fig.~\ref{fig:laplace}a, where the length of the arrows corresponds to the magnitude of the deformation. Next, with the identified point correspondence and deformation field, each domain $\Omegathetai$ can be parameterized using truncating principal component coefficients in PCA, yielding the low-dimensional shape parameter $\tilde{\theta}$.
Although the diffeomorphisms in this example are explicitly known, we intentionally use PCA to parameterize the shape to demonstrate the generalization of our framework when the diffeomorphisms are unknown. 
A comparison of network performance using $\theta$ and $\tilde{\theta}$ can be found in Sec.~\ref{append:pca_true}.

\noindent{\bf{Network training and performance}}. Fig.~\ref{fig:laplace}b illustrates the network architecture for approximating the latent operator on the reference domain for this example. Since the new operator is a function of shape parameters and boundary conditions, the network has two branch networks for inputting the truncated shape parameter $\tilde{\theta}$ and boundary values represented by a set of discrete points, denoted as $\tilde v$. The trunk network takes the coordinates on the reference shape ${\mathbf{x}}$ as input. We train the network on $n=3,300$ training cases (see \ref{append:net_detail} for details) and use the remaining $n=200$ cases for testing. Network predictions in $\Omegaref$ are then mapped back to the original domain via $\varphi_\theta$ for further visualization and error analysis.
Fig.~\ref{fig:laplace}c displays network predictions $u_{NN}$ and the absolute errors $|u_{FEM} - u_{NN}|$ on three representative domains in the testing cases, indicating that network predictions are in good agreement with the numerical solutions. The maximum value of the absolute error is less than $2e-2$. We summarize the statistics of the mean absolute error $\varepsilon_{abs}$ and relative $L_2$ error $\varepsilon_{l2}:=||u_{FEM}- u_{NN}||_2/||u_{FEM}||_2$ in the testing cases by presenting their distributions in Fig.~\ref{fig:laplace}d. Both distributions are left-skewed when viewed in a logarithmic scale, with the averaged $\varepsilon_{abs}$ and the mean relative $L^{2}$ error at $4.3\times 10^{-3}$ and $8.3\times10^{-3}$, respectively. 
The behavior of train and test error as a function of the number of training cases is represented in Fig.~\ref{fig:laplace}e. $\bar \varepsilon_{l2}$ of the testing dataset decreases to the same order of training error with an increasing number of training cases, from $\mathbf{O}(10^{0})$ to $\mathbf{O}(10^{2})$. Based on the above results, we conclude that our framework enables effective learning of the solution operator for the Laplace equation on a diffeomorphic family of domains. 

\subsection{Example 2: Learning Reaction-Diffusion Dynamics on Parametric 2D Domains}
\label{subsec:rd}

\begin{figure}
    \centering
	\includegraphics[width=1.0\textwidth]{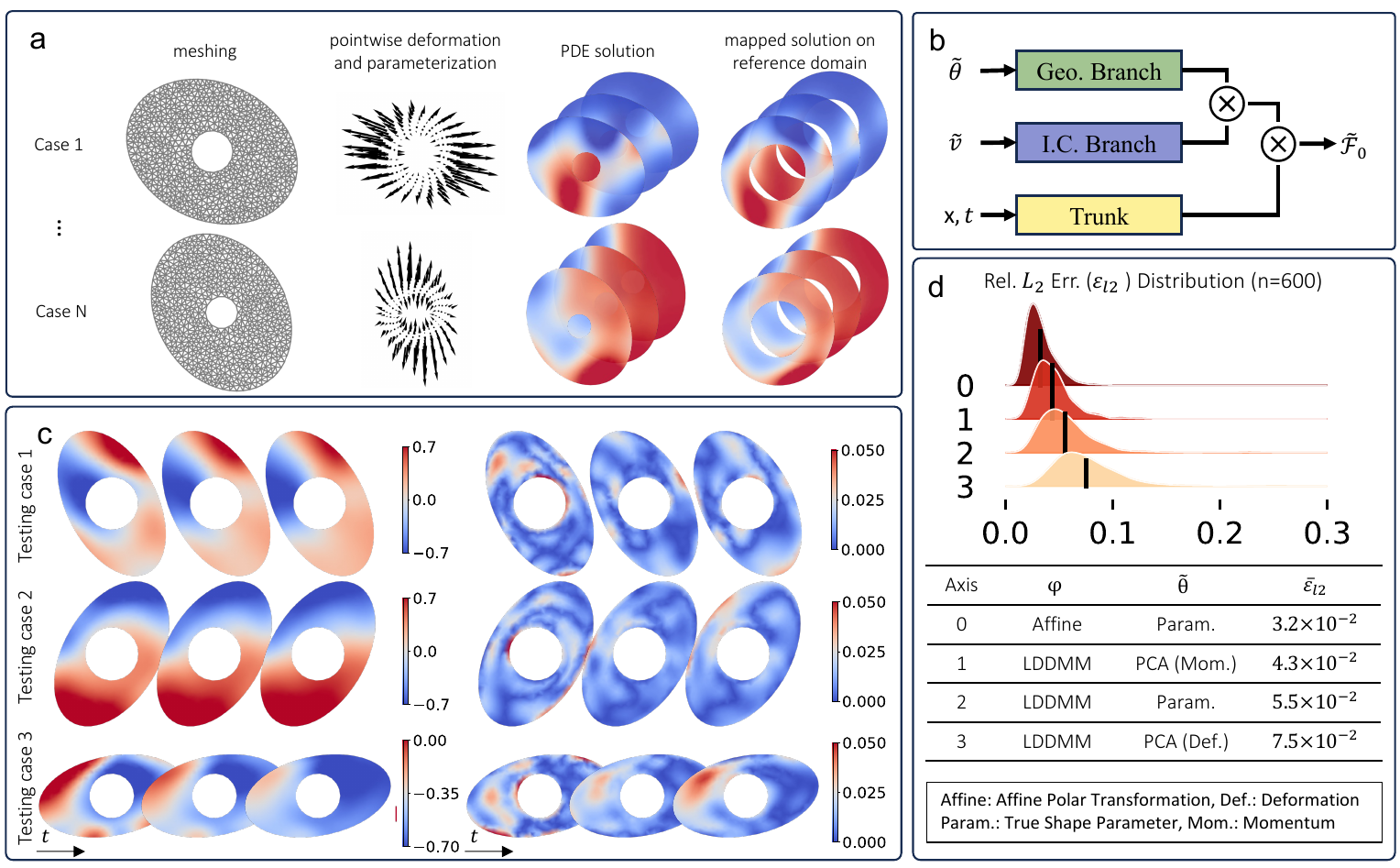} 
    \caption{\textbf{DIMON for learning solution operator of the Reaction-Diffusion (RD) equation on parametric 2D domains.} {\bf a}: Data preparation on $N$ parametric domains. Each domain is parameterized by either the deformation or the momentum field at the landmark points with $p'$ principal component coefficients. A finite element method is adopted to numerically solve the RD equation with various initial conditions (I.C.) on $\Omegathetai$, set as an annulus, and map the solution to $\Omegaref$ at different time snapshots. {\bf b}: Neural operator structure. Truncated shape parameter $\tilde \theta$ and boundary condition $\tilde v$ are fed into the geometry (Geo.) and boundary condition (B.C.) branch, respectively. Coordinates in the reference domain $\mathbf{x}$ are the input of the trunk net. {\bf c}: Network predictions with true shape parameters of the affine transformation at $t = 0, 1,$ and $2$s and their absolute errors on three testing domains. {\bf d}: Average relative $L_{2}$ error ($\varepsilon_{l2}$) on the testing cases based on $\tilde \theta$ obtained from 4 approaches: affine with true shape parameters, LDDMM with PCA on momentum and deformation, and LDDMM with true shape parameters. The averaged $\varepsilon_{l2}$ for different approaches range from $3.2\times10^{-2}$ to $7.5\times 10^{-2}$, indicating robustness of DIMON with respect to the way the diffeomorphisms are learned and approximated. The number of testing cases is $n=600$.}
    \label{fig:rd}
\end{figure}

\noindent{\bf{PDE problem}}. We now consider a more challenging example, with a PDE that includes time and is nonlinear, namely a reaction-diffusion in the form
\begin{align*}
    \frac{\partial u}{\partial t} &= \Delta u + R(u)\,,\quad \text{ with } R(u) = u(1-u^{2})\,,\quad \text{ in } \Omegatheta\\    
    u|_{t=0} &=v \,,\qquad\left.\frac{\partial u}{\partial n}\right|_{\partial\Omegatheta} = 0\,.
\end{align*}
where the $\Omegatheta$s are elongated ellipses with a central hole, and $\Omegaref$ is a circular annulus. 
The reaction term $R(u)$ is adopted from the Newell–Whitehead-Segel model~\cite{newell1969finite} to describe post-critical Rayleigh-B\'enard convection. We impose a Neumann boundary condition and impose initial conditions by sampling from a Gaussian Random Field (GRF). 

\noindent{\bf{Data preparation}}. We visualize the data generation and learning process in Fig.~\ref{fig:rd}a: we generate $N=6,000$ sets of shape parameters $\theta$ following the steps in Method and Materials~\ref{subsec:dom_gen}. Next, we simulate the system dynamics from $t=0$ to $t=2$ with a finite element method and map the numerical solution at different snapshots to the reference domain. 

\noindent{\bf{Construction of $\varphi_\theta$ and $\tilde{\theta}$}}. To demonstrate the compatibility of our framework with generic methods of domain mapping, we adopt two approaches for acquiring point correspondences between $\Omegathetai$ and $\Omegaref$ (Fig.\ref{fig:rd}a), namely, affine polar transformation (referred to as ``affine") and LDDMM with details provided in Sec.\ref{subsec:lddmm}. The former approach converts the Cartesian coordinates of $\Omegathetai$ into polar coordinates with the radius normalized to $[0.5, 1.0]$. However, affine transformation in the example is less general than LDDMM as it depends on the particular shape of $\Omegaref$ and $\Omegathetai$. With the identified point correspondence, we proceed to calculate the truncated shape parameters $\tilde \theta$ using PCA based on 1) deformation from LDDMM and 2) momentum from LDDMM.

\noindent{\bf{Network training and performance}}.
Fig.~\ref{fig:rd}b shows the architecture of the network. The network contains a branch for encoding the truncated shape parameter $\tilde{\theta}$ and one for the initial conditions $f$. $\mathbf{x}$ and $t$ are the coordinates in $\Omegaref$ and time.
We train a network with 5,400 training cases and examine its performance with 600 testing cases. Fig.~\ref{fig:rd}c displays the predicted solutions and the absolute errors for three representative testing cases at $t=0, 1,$ and $2$s. $\tilde \theta$ are adopted as the true shape parameters. Qualitatively, the network predictions closely match the finite element solutions with low absolute errors. Fig.~\ref{fig:rd}d illustrates the relative $L_2$ error distribution calculated over the 600 testing cases from various point correspondence algorithms, including 0) Affine with true parameter 1-3) LDDMM with PCA on momentum, true parameter, and PCA on deformation, where the true parameters refers to the shape parameters in Sec.~\ref{subsec:dom_gen}. All of the distributions exhibit left skewness with means (black line) ranging from $3.2\times10^{-2}$ to $7.5\times10^{-2}$. Mean errors from affine transformation are smaller than those of LDDMM due to the more uniform distribution of landmarks in $\Omegathetai$ and no matching error between $\Omegaref$ and $\Omegathetai$. A comparison of affine transformation and LDDMM is presented in~\ref{append:method_pt_corr}. Nevertheless, the above results indicate that DIMON can accurately learn the reaction-diffusion dynamics on the diffeomorphic family of domains with a flexible choice of point-matching algorithms.

\subsection{Example 3: Predicting patient-specific electrical wave propagation in the left ventricle}
\label{subsec:LV}

In this section, we demonstrate the capabilities of our framework on predicting electrical signal propagation on the left ventricles (LVs) reconstructed for clinical imaging scans of patients.

\noindent{\bf{PDE problem}}. 
Human ventricular electrophysiology is described by the following mono-domain reaction-diffusion equation:
\begin{equation}
    C_{m}u_{t} = -(I_{ion} + I_{stim}) + \nabla D \nabla u, 
\end{equation}
where $C_{m}$ is the membrane capacitance, $u$ is the transmembrane potential, $D$ is a conductivity tensor controlled by the local fiber orientation, $I_{stim}$ is the transmembrane current from external stimulation, and $I_{ion}$ is the sum of all ionic current. A comprehensive description of $I_{stim}$, $I_{ion}$, model parameters, and the underlying ordinary differential equations can be found in~\cite{ten2004model,ten2006alternans}. It is important to note that the system is essentially anisotropic and multiscale. 

\begin{figure}
    \centering
	\includegraphics[width=1.0\textwidth]{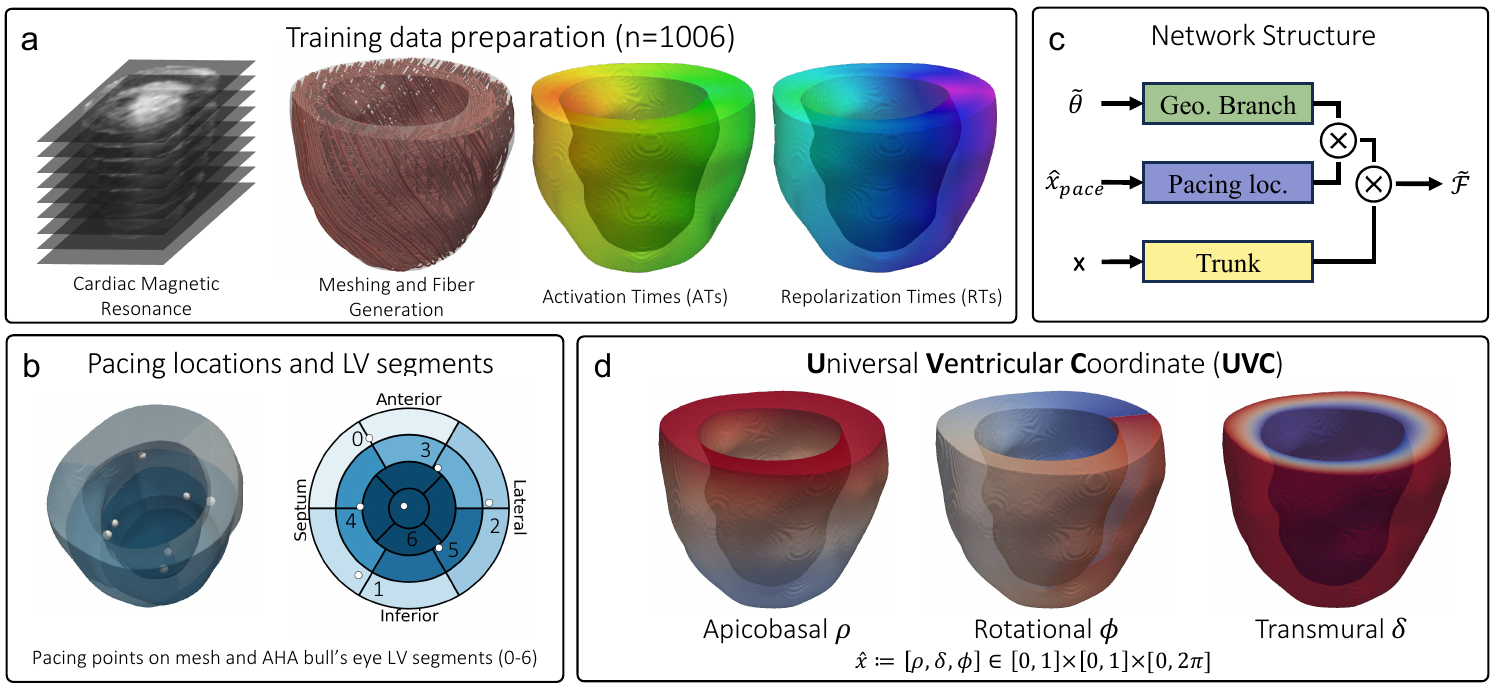} 
    \caption{\textbf{Predicting electrical signal propagation on patient-specific left ventricles (LV).} a: Data preparation for the network training. The cohort assembled consists of 1007 cardiac magnetic resonance (CMR) scans of patients from Johns Hopkins Hospital, based on which LV meshes and fiber fields are generated. Next, we simulate a paced beat for 600 ms at an endocardial site until the LV is fully repolarized and record the activation times (ATs) and repolarization times (RTs) for network training. b: Pacing locations (marked in white balls) located on the endocardium with a reduced American Heart Association (AHA) LV segments from zone 0 to 6. c: Network structure. c: Universal ventricular coordinate (UVC) calculation for registering points across geometries. }
    \label{fig:LV_data}
\end{figure}

\noindent{\bf{Data preparation}}. Fig.~\ref{fig:LV_data}a provides an overview of training data generation. We collected 1,007 anonymized cardiac magnetic resonance (CMR) scans from patients at Johns Hopkins Hospital. However, our goal in this work is not to predict the full solution of this multiscale system at all times; we focus on two clinically relevant quantities, the distributions of activation times (ATs) and repolarization times (RTs) of a propagating electrical signal, defined as the time when local myocardium tissue is depolarized (activated) and repolarized. For each LV, we apply an external electrical stimulation (pacing) at 7 myocardium points (Fig.~\ref{fig:LV_data}b) in each segment and record the local ATs and RTs for each elicited wave (visualized in Fig.~\ref{fig:LV_data}a) for network training and testing. We provide more details of data generation in Methods and Materials~\ref{subsec:cohort}.

\noindent{\bf{Construction of $\varphi_\theta$ and $\tilde{\theta}$}}. We choose a hollow semi-sphere with an inner radius of 0.5 and an outer radius of 1 as $\Omegaref$ for LVs as they are diffeomorphically equivalent. To calculate point correspondence between these non-rigid and unparameterized LV meshes with $\Omegaref$, we adopt a registration method by matching the universal ventricular coordinates (UVCs) ~\cite{bayer2012novel,bayer2018universal}. As depicted in Fig.~\ref{fig:LV_data}d, UVC converts Cartesian coordinates of a LV into three new coordinates in fixed ranges: apicobasal ($\rho \in [0, 1]$), rotational ($\phi \in [0, 2\pi]$), and transmural ($\delta \in [0, 1]$). Here, $\rho$ and $\delta$ represent a normalized ``distance" to the basal and epicardial surfaces, while $\phi$ represents a rotational angle with respect to the right ventricle center. We calculate the UVCs for each LV in the cohort and establish their correspondence to $\Omegaref$ by matching their UVCs. Once the point correspondence is attained, we perform PCA on the coordinate difference at landmark points to derive $\tilde{\theta}$, a low-dimensional representation of the shape parameter $\theta$. 

Notably, we assume the fiber field depends solely on the shape of the LV~\cite{bayer2012novel}, thereby encoding the LV shape with $\tilde{\theta}$ simultaneously leads to encoding the fiber field. To illustrate how the fiber fields influence electrical propagation, we present three exemplary cases with ATs on the reference domain in Sec.~\ref{app:map_sol}, each with the same pacing location.  

\begin{figure}
    \centering
	\includegraphics[width=1.0\textwidth]{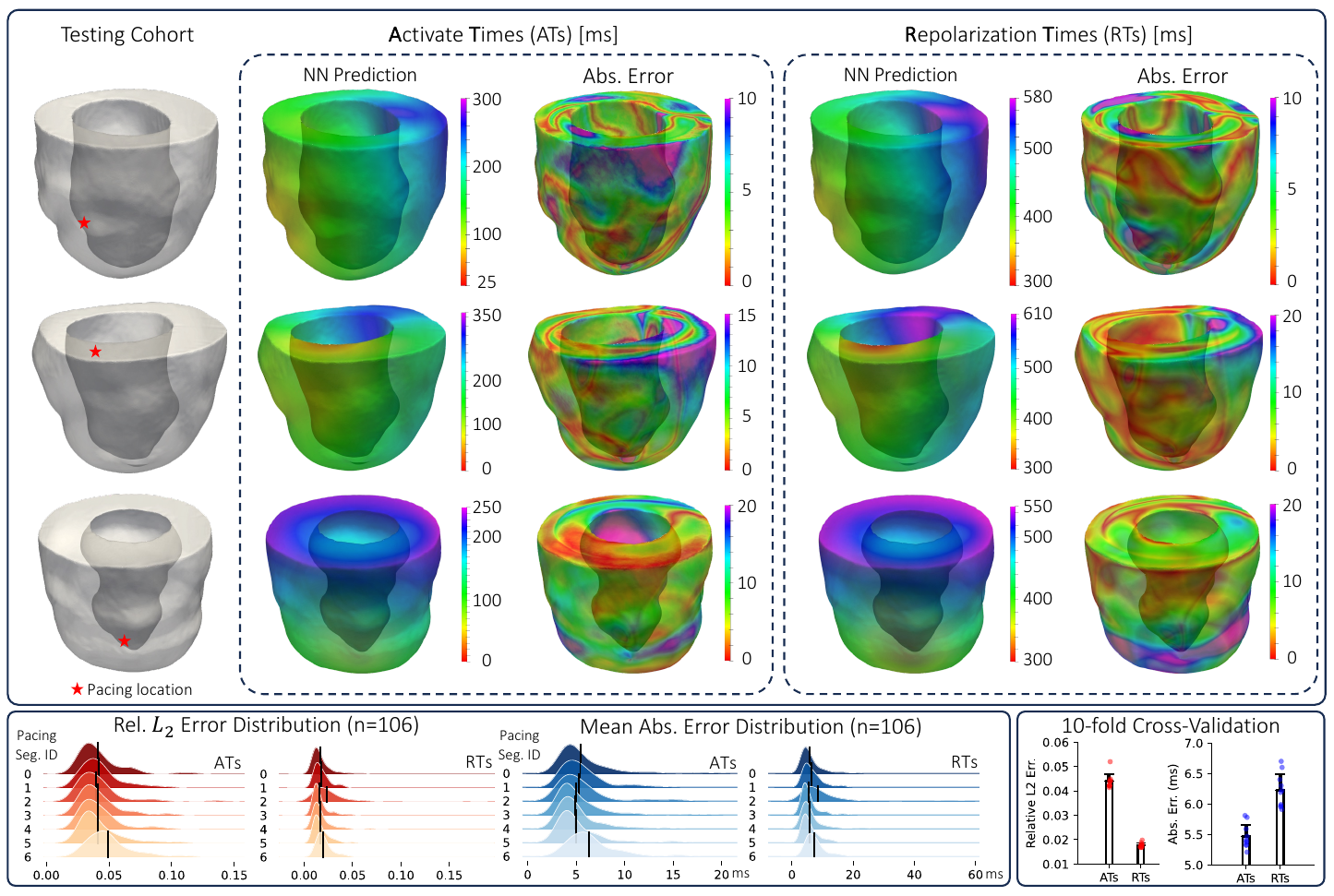} 
    \caption{\textbf{Network prediction of ATs and RTs from the testing cohort.} Top panel: LV shape with their predicted ATs and RTs and absolute errors. Red stars are the pacing locations. The colorbar is selected to be consistent with the CARTO clinical mapping system, with red denoting early activation and progressing to purple for late activation.  Bottom left panel: Relative and absolution error distributions for ATs and RTs. Bottom right panel: Bar plots of the average relative $L_2$ errors from 10-fold cross-validation.}
    \label{fig:LV_results}
\end{figure}

\noindent{\bf{Network training and performance}}. We study the neural network performance in predicting ATs and RTs on patient-specific LVs. The network architecture is presented in Fig.~\ref{fig:LV_data}b. We input shape parameters $\tilde{\theta}$ along with pacing locations $\hat{x}_{pace}$ into two separate branches. The trunk network takes UVCs ($\mathbf{x}$) of each training point as input. We present the network prediction in the top panel of Fig.~\ref{fig:LV_results}. After random shuffling, we split the training dataset (n=1007) into 10 subsets with 100 geometries in sets 1-9 and 107 geometries in set 10. First, we train a network using geometries and ATs/RTs in sets 1-9 and set 10 for testing. We visualize predicted ATs and RTs along with their absolute errors for three representative LVs from the testing cohort. We observe that ATs and RTs on each LV with various pacing locations are faithfully predicted by our framework, with the maximum absolute error of ATs and RTs being around 20 ms (purple areas in the ``Abs. Error'' columns in Fig.\ref{fig:LV_results}). This error is relatively small compared to the maximum values of ATs and RTs. 

To quantitatively investigate how well the network predicts for the testing cases, we present the distribution of relative $L_{2}$ and absolute error in the bottom panel. Each row represents a different myocardial segment as defined in Fig.~\ref{fig:LV_data}b. The error distributions are roughly centered around their respective means (black lines) with left skewness. However, for apical regions (segment 6), both errors are slightly larger than the rest of the regions, which could be attributed to the complex fiber architecture created by rule-based methods at the apex~\cite{doste2019rule}. More specifically, we found that for the three cases with the largest error, the reconstructed LV shape unfaithfully reflects the true shape of LV due to the collective artifacts from data acquisition and imaging segmentation. We discuss this further in Sec.~\ref{append:bad_shape}.

Furthermore, we perform 10-fold cross-validation to quantify the generalization ability of the network, e.g., adopting sets 2-10 for training and set 1 for testing. The bottom right panel shows the relative and absolute errors from 10-fold cross-validation along with their means (0.018 and 0.045) and standard deviations. In general, relative errors of ATs from 10 iterations are larger than the RTs errors, though the magnitude of errors in ATs is smaller than the RTs. 

Finally, we compare the computational cost between our neural operator approach and a traditional finite elements solver (openCARP, used widely by the cardiac computational modeling community) in Table~\ref{tab:comp_cost}. During the testing phase, our neural operator significantly accelerates the overall speed: it only requires less than 1 s on a single GPU to predict the ATs and RTs for a new patient, whereas it takes our solver openCARP 5 hours on 48 CPU nodes to simulate ATs and RTs in a single beat, not accounting for the additional time spent preprocessing and meshing. In conclusion, our framework demonstrates that a neural operator can efficiently predict patient-specific electric propagation on realistic heart geometries with satisfactory accuracy and substantial speedup in computing.

\begin{table}
    \centering
    \begin{tabular}{c  c  c}
        \hline
                    &  Wall time        & Hardware  \\
        \hline
        \textbf{Neural Operator}        &  \textbf{$<$1s}  & \textbf{1 GPU}       \\
        openCARP (FEM)   &  5 hours          & 48 CPU cores         \\
        \hline
    \end{tabular}
    \caption{\textbf{Computational costs of neural operator and openCARP for prediction.}}
    \label{tab:comp_cost}
\end{table}

\section{Discussion}
\label{sec:disc}
We introduced the DIMON framework for learning solution operators of PDEs on a family of diffeomorphic domains using neural networks. DIMON combines diffeomorphic maps between domains with a neural operator that learns a latent solution operator of PDEs, parameterized by the domain shape together with boundary and initial conditions. We first described the problem formulation, along with the unfolding of the solution of a pedagogical example. We then presented an extension of the approximation theorem for a neural operator to the setting introduced here that includes a diffeomorphic family of domains. Next, we discussed the overall workflow of DIMON, together with shape-matching algorithms for estimating in practice the diffeomorphisms between the shapes a reference shape. Finally, the performance of DIMON on operator learning is demonstrated by three representative examples, including solving static and dynamic problems on synthetic parameterized 2D domains or on realistic 3D geometries, including a data set of interest in cardiac electrophysiology and precision medicine. Networks were trained with various shape-matching algorithms and a compressed, low-dimensional shape parameter $\tilde \theta$, constructed using a variety of methods including LDDMM, affine transformation, or UVC coordinates, combined with the true shape parameter or PCA reduction on deformation/momentum fields. These examples demonstrate the capability of DIMON for efficient and accurate predictions of PDE solutions or functionals thereof on a variety of geometries.

DIMON is novel as it is the first systematic work that extends the learning ability of neural operators to PDEs on multiple geometries. We remove the restriction of learning on a fixed domain in \cite{jin2022mionet,lu2021learning} by introducing a generic diffeomorphic mapping to the framework and transporting functions in $\Omegatheta$ to a fixed domain $\Omegaref$. In addition, we also show that, by adopting neural operator architecture in~\cite{jin2022mionet} and introducing the shape space $\Theta$, DIMON is compatible with the approximation theorem described in~\cite{jin2022mionet}, which provides support for the theoretical development and practical implementation of our framework. 

The framework is quite general in many aspects. For example, although we chose MIONet as the baseline neural operator model in this work, DIMON is agnostic to the choice of neural operators, allowing to incorporate a variety of models, including~\cite{li2020fourier,lu2022multifidelity,cao2023lno,hao2023gnot}. We also demonstrate, in Example 3, that the learning ability of DIMON reaches beyond solution operators of PDEs. The ATs/RTs are the times when the PDE solution, transmembrane potential, exceeds or decreases below certain thresholds, and therefore functionals of solutions of the PDE, rather than solutions themselves. Finally, in the direction of design and optimization, DIMON allows for the computation of the derivative of the cost functional with respect to the parameters that describe the domain shapes~\cite{shukla2023deep} via automatic differentiation. This, in turn, simplifies the calculation of admissible deformations in shape optimization.
This framework conveniently allows imposing boundary conditions, or initial conditions, or local material properties on different geometries. There is also significant flexibility in selecting the diffeomorphic reference domain, and the algorithm to register $\Omegatheta$ to $\Omegaref$.

Our framework is significant to many downstream applications. One concrete example arises from cardiology. For instance, in cardiac electrophysiology, a novel approach to examine the arrhythmia inducibility of the substrate is to perform simulations based on the patient's heart shape acquired from high-quality imaging scans. Though effective~\cite{prakosa2018personalized}, the excessively high computational cost poses a significant challenge to its vast application and implementation in clinics. With our framework, one could train a neural operator that predicts electrical propagation on various hearts without the need to perform expensive numerical simulations. Our framework is also promising for topology optimization, where shapes are optimized in order to meet certain design requirements, and PDEs need to solved repeatedly on a large number of domains during such optimization procedure.

DIMON also has several limitations. We only consider mapping scalar functions in this work, whereas mapping vector or tensor fields is yet to be addressed given that it holds particular importance in representing stress and velocity in fluid and solid mechanics. Incorporating methods that transport vector/tensor fields across shapes or manifolds~\cite{beg2005computing,donati2022complex,azencot2013operator} can potentially address this issue and will be of interest in the future. Also, although PCA is indeed a general approach for parameterizing shapes irrespective of their topology, other local methods for shape parameterization, such as non-uniform rational B-spline (NURBS), can be adopted due to their explicit form of splines or polynomials~\cite{bucelli2021multipatch,willems2023isogeometric}. 
In addition, conducting a rigorous analysis of the network's convergence rate is beyond the scope of this study, making it difficult to estimate a priori how large of a training set is needed in order to achieve a given accuracy of the solutions.

\section{Material and Methods}

\subsection{Large Deformation Diffeomorphism Metric Mapping (LDDMM)}\label{subsec:lddmm}
We give a brief summary of the LDDMM framework that is used specifically to estimate diffeomorphic mappings between the domains of Example 2, but that could more generally apply to a variety of different situations and geometric objects. The interested reader can further refer to \cite{beg2005computing} and \cite{younes2019shapes} for additional details on the model. 

The central principle of LDDMM is to introduce diffeomorphic transformations of $\mathbb{R}^d$ which are obtained as the flows of time-dependent vector fields. More precisely, given a time-dependent integrable vector field $t \in [0,1] \mapsto v(t, \cdot)\in C^1_0(\mathbb{R}^d,\mathbb{R}^d)$ (the space of $C^1$ vector fields on $\mathbb{R}^d$ which vanish at infinity as well as their first derivatives). Then the flow map $\varphi^v$ of $v$, defined  by $\varphi^v(t,x) = x + \int_0^t v(s,\varphi^v(s,x)) ds$, can be shown to be a $C^1$-diffeomorphism of $\mathbb{R}^d$. The group of all such flows taken at the terminal time $t=1$ can be further equipped with a metric as follows. One first restricts to a given Hilbert space $V$ of vector fields in $C^1_0(\mathbb{R}^d,\mathbb{R}^d)$ (e.g. a Sobolev space of sufficient regularity or reproducing kernel Hilbert space with adequate kernel function) and define the energy of the flow map $\varphi^v(1,\cdot)$ by $\int_0^1 \|v(t,\cdot)\|_V^2 dt$. This provides a quantitative measure of deformation cost and, given two shapes $\Omega$ and $\Omega'$ in $\mathbb{R}^d$, an optimal transformation mapping $\Omega$ to $\Omega'$ can be found by introducing the registration problem:
\begin{equation}
\label{eq:LDDMM_matching}
\inf_{v \in L^2([0,1],V)} \left\{ \int_0^1 \|v(t,\cdot)\|_V^2 dt \ | \ \varphi^v(\Omega) = \Omega' \right\}
\end{equation}
In practice, it is common to further relax the boundary constraint $\varphi^v(\Omega) = \Omega'$ based on a similarity measure between the deformed domain $\varphi^v(\Omega)$ and $\Omega'$. This is typically done to ensure robustness of the method to noise and inconsistencies between the two shapes. 

Optimality conditions for the above problem can be derived from the Prontryagin maximum principle of optimal control (c.f. \cite{younes2019shapes} Chap. 10) from which it can be shown that a minimizing vector field for \eqref{eq:LDDMM_matching} can be parameterized by an initial costate (or momentum) variable. Furthermore, when $\Omega$ is a discrete shape (i.e. represented by a finite set of vertices), this initial momentum can be identified with a distribution of vectors over the vertices of $\Omega$. This allows to reduce the registration problem to a minimization over the finite-dimensional initial momentum variable, which can be tackled for instance with a shooting approach. For the applications presented in this paper, we make use specifically of the implementation of diffeomorphic registration from the MATLAB library \textit{FshapesTk}\footnote{https://plmlab.math.cnrs.fr/benjamin.charlier/fshapesTk}.

\subsection{Principal Component Analysis}\label{subsec:pca}
Consider $p$ landmark points on $\Omegaref$ with known point correspondences on each $\Omegathetai$ given $N$ domains, $A \in \mathbb{R}^{p\times N}$ denotes the Euclidean difference between the target domains and the reference domain in each direction. We define $\Bar{A} = A - \mu$, where $\Bar{A}$ is normalized to zero-mean. We perform the singular value decomposition (SVD) $\Bar{A} = U\Sigma V^{T}$, where $\Bar{A} \in \mathbb{R}^{N\times p}$, $U  \in \mathbb{R}^{N\times p}$, $\Sigma  \in \mathbb{R}^{p\times p}$, and $V^{T} \in\mathbb{R}^{p\times p}$. We truncate $U$ and $\Sigma$ to first $p'$ rank and result in reduced matrices as $U_{r} \in \mathbb{R}^{N\times p'}$, $\Sigma_{r}  \in \mathbb{R}^{p'\times p'}$, and $V^{T}_{r} \in\mathbb{R}^{p' \times p'}$. We define the coefficient matrix $D = U_{r}\Sigma_{r}$,
where each row in $D \in \mathbb{R}^{N\times p'}$ is a reduced representation of the deformation from $\Omegatheta$ to $\Omegaref$, referred to as reduced shape parameter $\tilde \theta$.

\subsection{Domain Generation}\label{subsec:dom_gen}
We generate the computational domain in Example 1 following the equations below. 
\begin{align*}
\begin{cases}
   x = X + a_{1}\cos{(a_{2}\pi + a_{3})}\sin{(a_{2}\pi + a_{3})}, \\
   y = Y + b_{1}\sin^{2}{(b_{2}\pi + b_{3})},\\
   a_{1},b_{1},a_{3}, b_{3} \sim U[-0.5, 0.5],\quad a_{2},b_{2} \sim U[0, 0.75],\\
\end{cases}
\end{align*}
where $X, Y \in [0,1]\times[0, 1]$ are the coordinates in the reference domain and $x,y$ are their corresponding coordinates on newly generated shape. $a_{i}, b_{j}$ are shape parameters, all chosen independently as follows: $a_1,a_3$ and $b_1,b_3$ from the uniform distribution in $[-0.5, 0.5]$ and $a_{2}, b_{2}$ from the uniform distribution on $[0,0.75]$.

In Example 2, we define the reference domain as an annulus with inner and outer radii at 0.5 and 1.0, respectively. The parametric family of domains is rotated ellipsis with a void. We sample the shape parameters following:
\begin{equation}
    c_{1} \sim U[1.5, 2.0], c_{2} \sim U[1.0, 1.5], c_{3} \sim U[0, \pi], c_{4} \sim U[0.2, 0.8], 
\end{equation}
Here, $c_{1}$ and $c_{2}$ stand for the length of the semi-major and minor axes, $c_{3}$ is the rotational angle, and $c_{4}$ is the radius of the inner circle. 

\subsection{Neural Operator} \label{subsec:NO}

We briefly summarize the architecture of MIONet in this section~\cite{jin2022mionet}. Notice that our framework is inclusive of the architecture of the neural operator we use to approximate $\mathcal{F}_{\refd}$. We denote $\tilde{\mathcal{F}}_{0}$ as an approximation of the mapping from the $m+1$ input functions $[\theta, v_{1},...,v_{m}]$ to the output function $u$. We remove the subscript that denotes the domain for clarity. We then define $m+1$ branch networks that input a discrete representation of each function and a trunk network that takes coordinates $\x  \in \Omegaref$ as input. Finally, the output function value at point $\x $ is approximated by
\begin{equation}\label{equ:no_general}
    \tilde{\mathcal{F}}_{0}(\theta,v_1,\cdots,v_m)(\x ) = \mathcal{S}\left( \underbrace{\tilde{\mathbf{g}}_0(\varphi_{q_0}^0(\theta))}_{\text{branch}_0} \odot \underbrace{\tilde{\mathbf{g}}_1(\varphi_{q_1}^1(v_1))}_{\text{branch}_1} \odot \cdots \odot \underbrace{\tilde{\mathbf{g}}_m(\varphi_{q_m}^m(v_m))}_{\text{branch}_m} \odot \underbrace{\tilde{\mathbf{f}}(\x )}_{\text{trunk}} \right),
\end{equation}
where $\tilde{\mathbf{g}}_i$ (called branch net $i$) and $\tilde{\mathbf{f}}$ (called trunk net) are neural networks, $\odot$ is the Hadamard product, and $\mathcal{S}$ is the summation of all the components of a vector~\cite{jin2022mionet}. Eq.~\eqref{equ:no_general} can be written in a simplified form as:
\begin{equation*}
    \tilde{\mathcal{F}}_{0}(\theta, v_{1,...,m})(\x ) = \sum^{k}_{i=1} t_{i}\prod^{m}_{j=1} b^{j}_{i}b^{geo}_{i} + b.
\end{equation*}
$b^{geo}_{i}$ is the output of Geo. branch given $\theta$, $t_{i}$ is the trunk network at $i$th neuron, and $b_{j}^{i}$ is the output of $i$th neuron from $j$th branch network. We refer the readers to~\cite{jin2022mionet,lu2021learning} for more details. In this work, we adopt the mean squared error (MSE) as the loss function. Higher-order constraints can be added to the network as further penalization~\cite{yin2022interfacing}. 

\subsection{Cohort and Simulations Details in Example 3}\label{subsec:cohort}
We used data from the Hopkins Hypertrophic Cardiomyopathy (HCM) Registry, which includes patients who were treated at the HCM Center of Excellence from 2005 to 2021. Our dataset consists of de-identiﬁed late gadolinium-enhanced cardiac magnetic resonance (LGE-CMR) scans from over 850 patients. The LGE-CMR scans have slice thicknesses ranging from $6-10$ $mm$ and spatial resolutions varying from $1.17$ to $2.56$ $mm$. Notice that some patients have multiple scans over time, which sums up to 1007 scans.

To simulate electrical propagation, we first segment the myocardium and the blood pool from the raw images. We then reconstruct a mesh and generate fibers for each LV based on the segmented images. The average length of mesh size is 400 microns, leading to the average number of vertices at the order of 4 millions~\cite{boyle2021characterizing,prakosa2018personalized}. To account for the directionality of the electrical signal propagated along the fibers, we adopt the validated rule-based method proposed in~\cite{bayer2012novel} to generate the myocardial fibers in each patient's LV. 

Next, we divide the myocardium surface into 7 segments, which is a simplified version of the standardized myocardium segmentation from American Heart Association (AHA)~\cite{american2002standardized,deng2019sensitivity}. Within each segment highlighted with a different color in Fig.~\ref{fig:LV_data}, we simulate a paced beat for 600 ms at a randomly-picked endocardial site until the LV is fully repolarized. Specifically, ATs and RTs are chosen as the time when the transmembrane potential $u$ reaches 0 mV from the resting state (-85 mV) and the time when $u$ decreases below -70 mV, recovering from depolarization. More details regarding the solver and simulation procedure can be found in~\cite{plank2021opencarp,prakosa2018personalized}.

\section{Acknowledgement} 
We would like to thank Dr. Zongyuan Li, Dr. Adityo Prakosa and Dr. Shane Loeffler for multiple discussions during the preparation of this manuscript. M.Y., R.B., N.T. are supported by grant NIH. M.Y. would like to acknowledge support from Heart Rhythm Society Fellowship. L.L. was supported by the U.S. Department of Energy [DE-SC0022953]. N.C. was supported by NSF grant DMS-1945224. M.M. is supported by AFOSR awards FA9550-20-1-0288, FA9550-21-1-0317, FA9550-23-1-0445.


\appendix
\section{Universal approximation theorem for the Laplace equation}
\label{append:laplace_case}
We provide here a more detailed explanation of the case of the Laplace equation discussed in Secs. \ref{subsec:prob_formul} and \ref{subsec:theorem}. We specifically focus on showing how the assumptions of Theorem \ref{thm:universal_approx} on the solution operator $\F0$ can be satisfied with adequate regularity conditions. We will restrict to the case where the reference domain $\Omegaref$ has a $C^2$ boundary (and thus so do all the domains $\Omegatheta$) and where boundary functions belong to the space $H^{3/2}(\partial \Omegaref)$. Let us then consider $\varphi \in C^2(\Omegaref,\mathbb{R}^d)$ a $C^2$ diffeomorphism from $\Omegaref$ to its image $\varphi(\Omegaref)$ as well as $g \in H^{3/2}(\partial \Omegaref)$. Following the discussion and notations in Sec. \ref{subsec:theorem}, we are interested in the solutions of the Dirichlet problem: 
\begin{equation*}
\begin{cases}
    \varphi^* \, \Delta \, (\varphi^{-1})^* \, u = 0 \qquad &\text{in }\Omegaref, \\
    u=g \qquad &\text{on } \partial\Omegaref
\end{cases}
\end{equation*}
which are the pullback of the solutions of the Laplace equation on the deformed domain $\varphi(\Omegaref)$. To be mathematically rigorous, the boundary condition needs to be taken in the sense of the trace on Sobolev spaces, that is, $T(u) = g$ where $T: H^{1}(\Omegaref) \rightarrow H^{1/2}(\partial \Omegaref)$ is the trace operator on $\Omegaref$. The operator $L_\varphi = \varphi^* \, \Delta \, (\varphi^{-1})^*$ can be expressed more explicitly with respect to the derivatives of $\varphi$ and takes the form:
\begin{equation*}
    L_\varphi u = \sum_{i,j=1}^{d} a^{ij}_\varphi(x) \, \partial_i \partial_j u + \sum_{i=1}^d b^i_\varphi(x) \, \partial_i u
\end{equation*}
and the coefficient functions are specifically given by:
\begin{equation*}
a^{ij}_\varphi(x) = \sum_{k=1}^d \partial_i \psi_k(\varphi(x)) \, \partial_j \psi_k(\varphi(x)) , \ \ b^i_\varphi(x) = \Delta \psi_i(\varphi(x))
\end{equation*}
where $\psi = \varphi^{-1}$ is a $C^2$ map from $\varphi(\Omegaref) \rightarrow \Omegaref$ and $\psi_j$ denotes the $j$-th coordinate of $\psi$. Given that both $\varphi$ and $\psi$ are $C^2$ maps, $a^{ij}_\varphi$ are $C^1$ functions and $b^i_\varphi$ continuous functions on $\Omegaref$. Furthermore, the strict ellipticity of the resulting operator $L_\varphi$ can be easily shown using the continuity of the derivatives of the inverse mapping $\psi$ on the compact domain $\varphi(\Omegaref)$. Since any boundary function $g \in H^{3/2}(\partial \Omegaref)$ is the trace of a function in $H^2(\Omegaref)$, it follows from \cite{gilbarg1977elliptic} (Theorems 8.3 and 8.12) that the Dirichlet problem has a unique weak solution $u_{\varphi,g}$ which further satisfies $u_{\varphi,g} \in H^2(\Omegaref)$ and the PDE $L_\varphi u_{\varphi,g}(x) = 0$ holds for almost all $x \in \Omegaref$.

In order to show the continuity of $u_{\varphi,g}$ with respect to $\varphi$ and $g$, we shall rely on an implicit function argument inspired from \cite{henry2005perturbation}. Denoting by $\text{Emb}^2(\Omegaref,\mathbb{R}^d)$ the space of $C^2$ embeddings from $\Omegaref$ to $\mathbb{R}^d$ equipped with the standard $\mathcal{C}^2$-norm, we introduce the following mapping: 
\begin{align*}
G:H^2(\Omegaref) \times \text{Emb}^2(\Omegaref,\mathbb{R}^d)  \times H^{3/2}(\partial \Omegaref) &\rightarrow L^2(\Omegaref) \times H^{3/2}(\partial \Omegaref) \\
(u,\varphi,g) &\mapsto (L_\varphi u, T(u)-g)
\end{align*}
Note that $u$ is a solution of the Dirichlet problem for $\varphi$ and $g$ if and only if $G(u,\varphi,g) = (0,0)$. Furthermore, as $L_\varphi$ and $T$ are continuous linear operators, $G$ is continuously Fr\'{e}chet differentiable with respect to $u$ and $g$. Moreover, with the assumptions made on the domain and on the functions regularity, it follows from the results of \cite{henry2005perturbation} Chap. 2 that $G$ is also continuously Fr\'{e}chet differentiable with respect to $\varphi$. Now, given any $(u,\varphi,g)$ such that $G(u,\varphi,g) = (0,0)$, we see that for any $v \in H^2(\Omegaref)$, $D_u G(u,\varphi,g) \cdot v = (L_\varphi v, T(v))$. Using again the existence and uniqueness of the solution to the Dirichlet problem stated in the previous paragraph (for $\varphi$ and the boundary function $T(v)$), we deduce that $D_u G(u,\varphi,g)$ induces an isomorphism between $H^2(\Omegaref)$ and $L^2(\Omegaref) \times H^{3/2}(\partial \Omegaref)$. The implicit function theorem then allows to conclude that, locally, the solution $u_{\varphi,g}$ can be expressed as a Fr\'{e}chet differentiable function of $\varphi$ and $g$. In particular, $u_{\varphi,g}$ is a continuous function of the embedding $\varphi$ and the boundary function $g \in H^{3/2}(\partial \Omegaref)$. 

Going back to the illustrative example of Sec. \ref{subsec:theorem}, as it is also assumed that $\theta \mapsto \varphi_\theta$ is continuous, the above analysis shows that the solution operator $\F0$ for that Dirichlet problem is indeed well-defined and continuous with respect to both the shape parameter $\theta \in \Theta$ and boundary condition $g^0 \in H^{3/2}(\Omegaref)$. Consequently, the universal approximation theorem applies to that operator restricted to $\Theta \times K$ where $K$ is any compact subset of $H^{3/2}(\partial \Omegaref)$. 

\section{Network training with a reduced and a true shape parameter}\label{append:pca_true}

To compare the network performance based on PCA coefficients and the true shape parameters ($a_{1,2,3}$ and $b_{1,2,3}$) in Example 1, we tabulate the mean absolute error and mean relative $L_{2}$ error in Table\ref{tab:pca_true}. For consistency, we retrain the network in Example 1 using the true parameters as input (See. Sec.~\ref{subsec:dom_gen}) and PCA coefficients truncated at $15$th modes while keeping other environments the same, such as hidden layers, activation functions, training epochs, or GPU machines. It is shown in Table~\ref{tab:pca_true} that network performance under two circumstances is almost the same, with the error using PCA being slightly lower than when training with the true shape parameters at 50,000 epochs. 

\begin{table}
    \centering
    \begin{tabular}{c  c  c}
        \hline
                    &  Abs. Err.        & Rel. Err.  \\
        \hline
        PCA coefficients ($n=15$)        &  $5.3\times 10^{-3}$  & $9.5\times 10^{-3}$      \\
        true shape parameter    &  $5.9\times 10^{-3}$         & $1.0\times 10^{-2}$        \\
        \hline
    \end{tabular}
    \caption{\textbf{Network prediction errors given the true shape parameters and PCA coefficients.} PCA is truncated at 15th modes.}
    \label{tab:pca_true}
\end{table}

\section{Affine Polar Transformation and LDDMM}\label{append:method_pt_corr}

\begin{figure}
    \centering
    \includegraphics[width=1.0\textwidth]{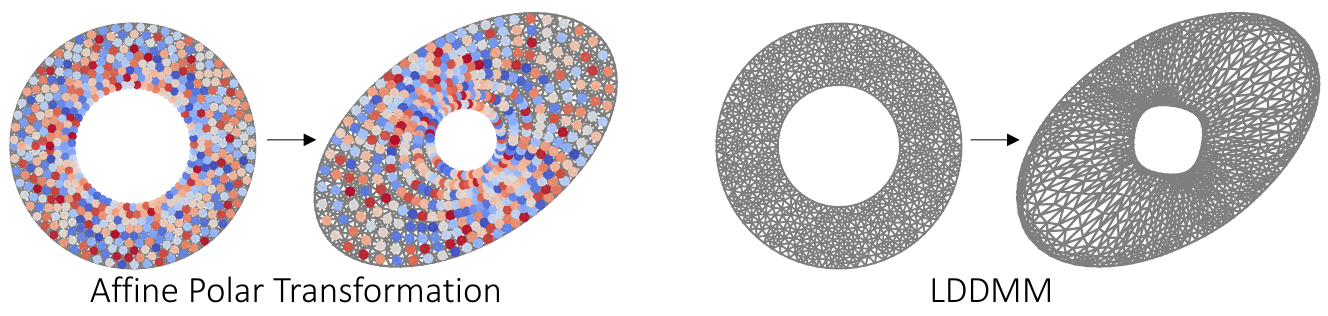} 
    \caption{\textbf{Affine Polar Transformation vs LDDMM.}}
    \label{fig:lddmm_aff}
\end{figure}

In Fig.~\ref{fig:lddmm_aff}, we present how landmarks are distributed on $\Omegaref$ and $\Omegathetai$ in Example 2. Affine polar transformation, as a special approach to match landmarks distributed on two shapes, has a more uniform point distribution. However, this approach lacks generality and can not be applied to general meshes. LDDMM, as a more general shape-matching algorithm, can register points distributed on arbitrary diffeomorphic shapes by morphing the mesh of $\Omegaref$ to approximate $\Omegathetai$. Yet, given its generality, it approximates the target shape with an error, displayed by the non-circular shape on the right panel of Fig.~\ref{fig:lddmm_aff}, and typically leads to a less uniform distribution of landmarks. These two reasons influence the accuracy of the encoding of the shape and of the sampling of training data, which are likely the case of the slightly larger testing error when using LDDMM than when using affine transformations.

\section{Network details} \label{append:net_detail}


We present network structure and training details in this section. Table~\ref{tab:network} shows the network size, training epochs, and training time on a single GPU for each example. We only adopt feedforward neural networks (FNNs) for all subnetworks. Let us take Example 1 for instance, the input dimension of the Geo. branch in Example 1 is 20 with 2 hidden layers. All the other subnetworks have two or three hidden layers. The nonlinear activation function is a hyperbolic tangent function. The number of training epochs ranges from 10,000 to 200,000 for each example, depending on the complexity and the size of the training datasets. All training processes are completed on an NVIDIA GeForce RTX 2060 GPU. Training the network in Example 1 and 3 costs less than 0.3 hours and the training in Example 2 costs 8.1 hours. Notably in Example 3, we expand the 1D rotational angle $\theta$ into a 2D vector, $[\cos{\theta}, \sin{\theta}]$, to impose the periodicity of the ATs/RTs at $\theta=0$ and $2\pi$. 

\begin{table}
    \centering
    \begin{tabular}{c  c  c  c  c  c}
        \hline
                    & Branch 1 (Geo) & Branch 2 & Trunk & Epochs & Training Time (h)\\
                    \hline
        Ex. 1   & [20, 100, 100, 100] & [68, 150, 150, 150, 100] & [2, 100, 100, 100] & 50,000 & 0.3 \\
        Ex. 2   & [75, 300, 300, 200] & [50, 200, 200, 200, 200] & [2, 100, 100, 100] & 100,000 & 8.1\\
        Ex. 3   & [180, 250, 250, 250, 200] & [3, 200, 200, 200, 200] & [4, 200, 200, 200, 200] & 10,000 & 0.17\\
                    \hline
    \end{tabular}
    \caption{\textbf{Network size, training epochs, and training time.}}
    \label{tab:network}
\end{table}

\section{Different PDE solutions on the same canonical shape}\label{app:map_sol}
\begin{figure}
    \centering
	\includegraphics[width=1.0\textwidth]{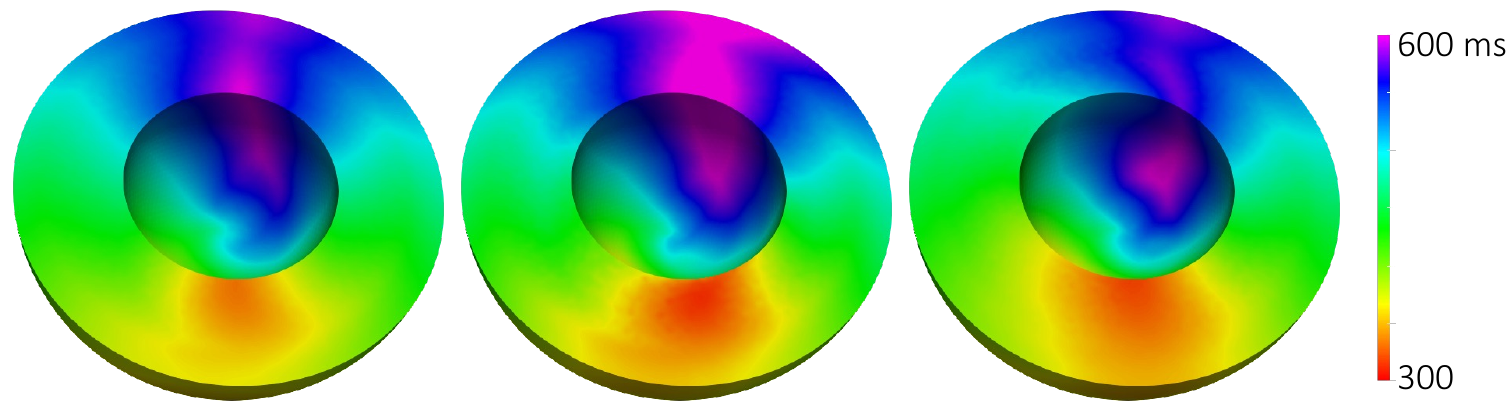} 
    \caption{\textbf{RTs mapped in the canonical domain with stimulation at the same endocardial point.}}
    \label{fig:mapped_RTs}
\end{figure}

In Fig.~\ref{fig:mapped_RTs}, we conduct simulations and display RTs from three distinct LVs on the canonical domain. For these three simulations, we keep all other variables, such as model parameters and pacing location, constant. Although the shapes of these LVs are distinctive, the UVCs of the pacing locations are the same. Thereby, any differences in RTs are solely induced by the variations in geometry (or ``fibers"). Myocardium tissue is stimulated on one side of the endocardium, where electric signals travel around the LV and eventually converge on the side of the LV. In the left panel, we observe that the tissue around the pacing location takes longer to repolarize compared to the other LVs. It takes a long period for the middle geometry to return to the repolarization state. This is indicated by a larger region of purple color, representing a longer repolarization time. These simulations vividly illustrate how the propagation of cardiac electrical signals is influenced by the geometry of the LV.

\section{Shape Outliers} \label{append:bad_shape}

\begin{figure}
    \centering
	\includegraphics[width=1.0\textwidth]{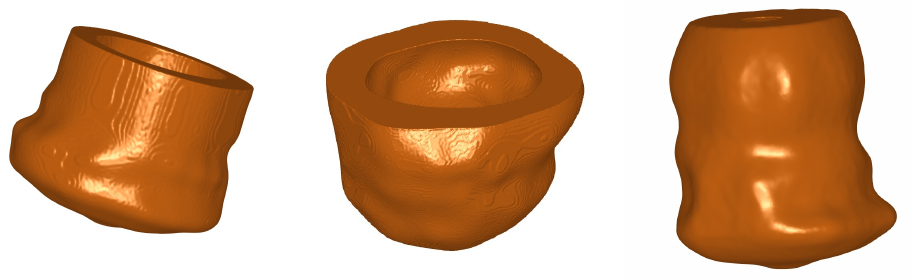} 
    \caption{\textbf{Bad shape reconstruction leads to larger errors that are presented in Fig.~\ref{fig:LV_results}}}
    \label{fig:outlier_shape}
\end{figure}

Fig.~\ref{fig:outlier_shape} presents three LV shapes from the case with the largest relative error in Fig.~\ref{fig:LV_results}. It is evident that these shapes poorly reflect the true shape of the LV, which can be attributed to three main reasons: poor segmentation, low-quality data, and misaligned imaging. The deep learning algorithms used for segmentation may erroneously distinguish the myocardium from the blood pool, leading to poor shape reconstruction~\cite{lefebvre2022lassnet}. The issue is exacerbated when segmenting low-quality images. Also, we noticed that the MRI scans for a few patients are not aligned well due to patient's breathing or inconsistent cardiac cycle during data collection. 

Calculating the distance between these outlier shapes to other shapes can help us to quantitatively understand the learning performance as a function of data distribution. Different metrics, including those used in large deformation diffeomorphic metric mapping (LDDMM), or Kendall's distance, can be adopted to quantify the distance~\cite{beg2005computing,hartman2023elastic}. The influence of the distance between shapes on learning performance of DIMON is an interesting problem, beyond the scope of this paper.

\bibliographystyle{abme}
\bibliography{reference}
\end{document}